\documentclass[journal]{IEEEtran}
\usepackage{graphicx}
\usepackage{cite}
\usepackage{amsmath}
\usepackage{multirow}
\usepackage{pbox}
\usepackage{subfig}
\usepackage{dblfloatfix}
\usepackage{hyperref}
\hypersetup{
    colorlinks=true,
    linkcolor=blue,
    filecolor=blue,      
    urlcolor=blue,
    citecolor=blue,
}
\usepackage{cleveref}

\renewcommand{\baselinestretch}{0.975}

\begin{document}
\title{Experimental System Identification and Disturbance Observer-based Control for a Monolithic $\mathrm{Z\theta_{x}\theta_{y}}$ Precision Positioning System}

\author{
	\vskip 1em
	{Mohammadali Ghafarian$^{1,2}$, %~\IEEEmembership{Member,~IEEE,} 
	Bijan Shirinzadeh$^{1}$, %~\IEEEmembership{Member,~IEEE,} 
	Ammar Al-Jodah$^{1,3}$, %~\IEEEmembership{Member,~IEEE,} 
	Tilok Kumar Das$^{1}$, %~\IEEEmembership{Member,~IEEE,}
	Tianyao Shen$^{1}$ %~\IEEEmembership{Member,~IEEE,}  
	}
	
	\thanks{$^{1}$Robotics and Mechatronics Research Laboratory (RMRL), Department of Mechanical and Aerospace Engineering, Monash University, Clayton, VIC 3800, Australia. $^{2}$Institute for Intelligent Systems Research and Innovation (IISRI), Deakin University, Geelong Waurn Ponds, VIC 3216, Australia. $^{3}$The University of Western Australia, Perth, WA 6009, Australia.}
	\thanks{Corresponding author: m.ghafarian@deakin.edu.au} 

}

%\markboth{IEEE/ASME Transactions on Mechatronics}%

\maketitle
	
\begin{abstract}
A compliant parallel micromanipulator is a mechanism in which the moving platform
is connected to the base through a number of flexural components.
Utilizing parallel-kinematics configurations and flexure joints,
the monolithic micromanipulators can achieve extremely high motion resolution and
accuracy. In this work, the focus was towards the experimental evaluation of
a 3-DOF ($\mathrm{Z\theta_{x}\theta_{y}}$) monolithic flexure-based piezo-driven micromanipulator for precise out-of-plane micro/nano positioning applications. The monolithic structure avoids the deficiencies of non-monolithic designs such as backlash, wear, friction, and improves the performance of micromanipulator in terms of high resolution, accuracy, and repeatability. A computational study was conducted to investigate and obtain the inverse kinematics of the proposed micromanipulator. As a result of computational analysis, the developed prototype of the micromanipulator is capable of executing large motion range of $\pm$238.5$\mathrm{\mu m}$ $\times$ $\pm$4830.5$\mathrm{\mu rad}$ $\times$ $\pm$5486.2$\mathrm{\mu rad}$. Finally, a sliding mode control strategy with nonlinear disturbance observer (SMC-NDO) was designed and implemented on the proposed micromanipulator to obtain system behaviours during experiments. The obtained results from different experimental tests validated the fine micromanipulator's positioning ability and the efficiency of the control methodology for precise micro/nano manipulation applications. The proposed micromanipulator achieved very fine spatial and rotational resolutions of $\pm$4$\mathrm{nm}$, $\pm$250$\mathrm{nrad}$, and $\pm$230$\mathrm{nrad}$ throughout its workspace.

\end{abstract}

\def\abstractname{Note to Practitioners}
\begin{abstract}
Piezo-actuated precision positioning systems play an increasingly important role in the fields of micro/nano manipulation robots. They have the advantages of fine resolution, high accuracy, fast response speed, and large output displacement. However, such systems inherently exhibit vibration, hysteresis behaviors, and are affected by external disturbances that could cause oscillations and positioning errors. This study presents a robust control methodology implemented on a 3-DOF positioning system ($\mathrm{Z\theta_{x}\theta_{y}}$), which is among the most prone system to be affected by existing disturbances. This control methodology is designed to improve the tracking performance in the presence of hysteresis nonlinearity, disturbances, and modeling errors. The effectiveness of the proposed control methodology is demonstrated by conducting a series of experiments. Due to the ease of implementation, the developed control methodology can be applied to other positioning systems as well.

\end{abstract}


%\markboth{IEEE TRANSACTIONS ON INDUSTRIAL ELECTRONICS}%
%{}

\vspace{-3mm}

\section{Introduction}
\IEEEPARstart{F}lexure joints have dominant superiority over traditional mechanical joints in precision engineering including micromanipulation mechanisms. Flexure-based parallel micromanipulators benefit from the advantages of both flexure joints and parallel-kinematics configurations, and additionally utilize the important characteristics of micromanipulators such as frictionless motion, absence of mechanical play and backlash, and no need for lubrication. These features are important and effective for various micro/nano positioning and nano-alignment applications, and it is not surprising that flexure-based parallel micro/nano manipulation systems stand out among others as the key element in the ultra-precision technologies \cite{GH2018,das2020characterization,al2018design,Yang2017a,Al-Jodah2020,Ghafarian2019,ALJODAH2021104334,MGhafarian_2020,Ghafarian_2020sms,das2018flexure}. Wei and Xu \cite{Wei2020} proposed a force-sensing cell microinjector based on a single-axis compliant small-stiffness mechanism. Yong et al. \cite{Yong2016a} presented single- and dual-stage vertical positioners for high-speed piezoelectric nanopositioning applications. Wang et al. \cite{Wang2019} presented a decoupled piezo-driven XY micro/nano positioning system with a travel range of $\pm27.7\times\pm26.6\mathrm{\mu m^2}$. Tian et al. \cite{Tian2019} presented a custom made atomic force microscopy (AFM) which was integrated with a 3-DOF XYZ parallel-kinematics piezo-driven micromanipulator for high-speed imaging. Cai et al. \cite{Cai2017a} presented the design and analysis of two parallel compliant piezo-driven 3-DOF micro/nano positioning system. Meanwhile, the inverse Bouc-Wen (BW) model was applied as a feedforward hysteresis compensator in the feedforward/feedback hybrid control method to compensate for the hysteresis of piezoelectric actuators (PEAs). The proposed micromanipulator exhibited small translational and rotational workspaces of $\pm4.1\times\pm5.2\times\pm6.5\mathrm{\mu m^3}$ and $\pm112\times\pm52.5\times\pm48.5\mathrm{\mu rad^3}$, respectively. Qin et al. \cite{Qin2013} proposed the design of a 3-PRR (prismatic-revolute-revolute) $\mathrm{XY\theta}$ micromanipulator. The Scott-Russell (SR) mechanism was utilized in the proposed design to magnify the displacement of the PEA. Dong et al. \cite{Dong2015} presented a 6-DOF piezo-driven micromanipulator. As a result of having no amplification mechanism, the workspace of the micromanipulator was very limited. Yang et al. \cite{Yang2017b} demonstrated the design, modeling, and experimental analysis of a piezo-driven $\mathrm{XY}$ micromanipulator. The experimental results illustrated that the $\mathrm{XY}$ micromanipulator had a working range of $\pm75\times \pm73.5\mathrm{\mu m^2}$ with the resolution of $\pm0.128\mathrm{\mu m}$ and $\pm0.143\mathrm{\mu m}$ in the X- and Y-directions, respectively.\\\begin{figure}[!t]\centering
	\begin{tabular}{c}
		\hbox{\hspace{-1em}\includegraphics[width=9cm]{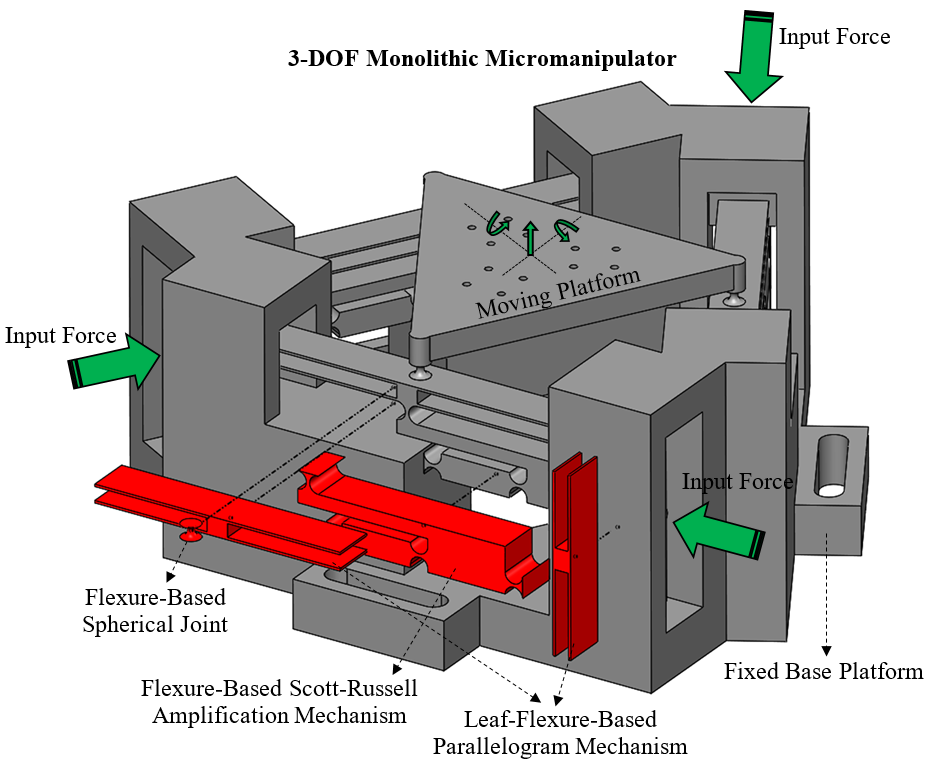}} \\ [0ex]
	\end{tabular}
	\caption{Monolithic $\mathrm{Z\theta_x\theta_y}$ parallel micromanipulator}\label{FIG_1}
	\vspace{-2mm}
\end{figure}Among the flexure-based parallel micromanipulators, the compliant $\mathrm{Z\theta_{x}\theta_{y}}$ type micro/nano positioning systems have become the research focus due to their important advantages for in out-of-plane positioning tasks \cite{Kim2013a,Lee2013,Kim2009,Pham_2019}. Qu et al. \cite{Qu2016} presented the design, modeling and test of a piezo-driven $\mathrm{\theta_{x}\theta_{y}}$ flexure-based micro/nano positioning system. The experimental results indicated that the developed micromanipulator could achieve a workspace of $\pm515\mathrm{\mu rad} \times \pm460\mathrm{\mu rad}$ about its two working axes with a resolution of $\pm0.5\mathrm{\mu rad}$. Chen et al. \cite{Chen2018a} introduced the mechanical design, modeling and experimental tests of a large-angle $\mathrm{Z\theta_{x}\theta_{y}}$ macromanipulator driven by four small air-gap voice coil actuators. The proposed system could achieve rotational-motion ranges of $\pm41.59\mathrm{mrad}$ and $\pm41.13\mathrm{mrad}$ in the working axes, for which the $\mathrm{Z}$-mode frequency was $49.6\mathrm{Hz}$ and the rotational ones were $55.45\mathrm{Hz}$ and $56.09\mathrm{Hz}$, respectively. The motion resolution of the macromanipulator was $\pm6.67\mathrm{\mu rad}$. Kim et al. \cite{Kim2010} presented an active vibration control system which was constructed based on a non-monolithic 3-DOF $\mathrm{Z\theta_x\theta_y}$ micro/nano manipulation system with an in-plane dimension of $160\mathrm{mm}$ (diameter) and an out-of-plane height of $60\mathrm{mm}$. Cao and Chen \cite {Cao2013} demonstrated the development of a system identification model for a commercially-available 3-DOF piezo-driven $\mathrm{Z\theta_{x}\theta_{y}}$ micromanipulator (P-558.TCD, Physik Instrumente). The system was driven by four PEAs and had a motion range of $\pm 25\mathrm{\mu m} \times \pm 250\mathrm{\mu rad} \times \pm 250\mathrm{\mu rad}$. The quasi-static analysis of a non-monolithic compliant tripod system for micro/nano positioning applications was presented by Wei et al. \cite{Huaxian2017}. The proposed micromanipulator had an overall positioning range of $\pm41\mathrm{\mu m} \times\pm330\mathrm{\mu rad}\times\pm385\mathrm{\mu rad}$. Considering the above-mentioned studies, a compliant monolithic 3-DOF piezo-driven micromanipulator was introduced by the authors with a larger workspace and fine resolution capable of executing three out-of-plane motions, one translation and two rotations. The structure of the proposed monolithic micromanipulator was optimized completely to have a maximum working range and bandwidth frequency higher than 100Hz \cite{Ghafarian2020}.\\
Regardless of the types of micromanipulators used to perform micro/nano manipulation tasks, an effective motion tracking control strategy can improve the tracking performances of the system significantly. In addition, disturbances such as cross-coupling, parametric uncertainties, etc. can practically affect and degrade the performance of a precision positioning system. Therefore, designing and utilizing a disturbance observer-based control methodology to be able to estimate and compensate the effect of disturbances for achieving high precision applications is very beneficial \cite{Pi2010,Zhang2016a,Zhang2016}. Chen et al. \cite{Chen2000} introduced a nonlinear disturbance observer (NDO) for robotic manipulators for various purposes such as friction compensation, independent joint control, sensorless torque control, and fault diagnosis. Furthermore, the global exponential stability of the proposed NDO was guaranteed. Lau et al. \cite{Lau2019} presented an enhanced adaptive robust disturbance observer-based motion tracking control methodology for tracking a desired motion trajectory in the presence of unknown or uncertain system's parameters, nonlinearities including hysteresis, and disturbances in the motion system. The proposed control methodology was applied in a semi-automated hand-held ear surgical device for the treatment of Otitis Media with Effusion (OME).\\\begin{table}[!t]
	\renewcommand{\arraystretch}{1.3}
	\caption{Mechanical and physical properties of ABS}
	\centering
	\label{table_1}
	%\centering
	\resizebox{\columnwidth}{!}{
		\begin{tabular}{l l l}
			\hline\hline \\[-3mm]
			\multicolumn{1}{l}{\textbf{Symbol}} & \multicolumn{1}{l}{\textbf{Quantity}} & \multicolumn{1}{l}{\pbox{20cm}{\textbf{Value}}}  \\[0ex] \hline
			\pbox{20cm}{$ \mathrm{\nu} $} & \pbox{20cm}{Poisson's ratio} &  \pbox{20cm}{$ 0.35 $ } \\ 
			$ \mathrm{\rho} $  & Density & $ 0.9087 \mathrm{(g/cm^3)} $ \\[0ex]
			$ \mathrm{E} $ & Young's modulus & $ 2200 \mathrm{(MPa)} $ \\[0ex]
			\pbox{20cm}{$ \mathrm{\sigma_{yield}} $ } & \pbox{20cm}{Tensile yield strength } & \pbox{20cm}{$ 31 \mathrm{(MPa)} $ } \\ [0ex]
			\pbox{20cm}{$ \mathrm{\sigma_{ultimate}} $ \\ \hphantom{1}} & \pbox{20cm}{Tensile ultimate strength } & \pbox{20cm}{$ 55 \mathrm{(MPa)} $ } \\  [0ex]
			\hline\hline
		\end{tabular}
	}
	\vspace{-1mm}
\end{table}\begin{figure*}[!t]\centering
	\begin{tabular}{c}
		\hbox{\hspace{0em}\includegraphics[width=16cm]{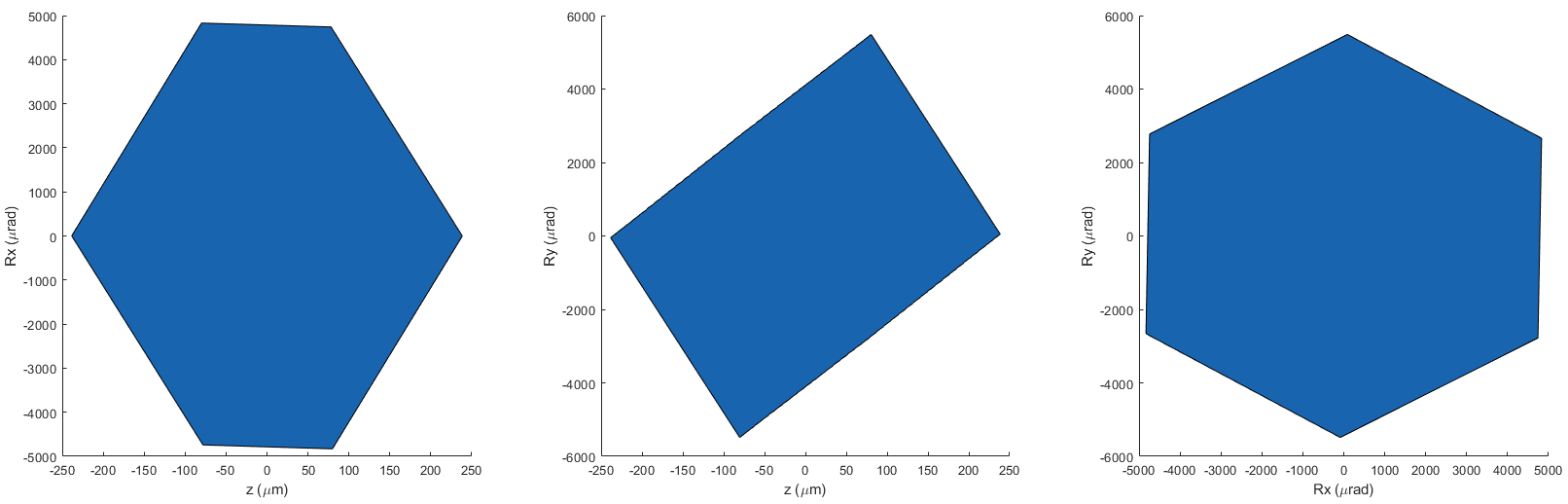}} \\ [0ex]
	\end{tabular}
	\caption{Reachable workspace of the developed micromanipulator}\label{FIG_2}
	\vspace{-5mm}
\end{figure*}Motivated by the previous work \cite{Ghafarian2020}, an experimental study of a large range piezo-driven spatial compliant monolithic parallel $\mathrm{Z\theta_x\theta_y}$ micro/nano manipulation system with a fine resolution is presented in this paper. Monolithically manufactured designs are very important for micro/nano applications and they are preferable in comparison with the assembled manipulation structures because of elimination of the unwanted features that affect a smooth and accurate nano-resolution manipulation. Other advantages of the proposed micromanipulator are low manufacturing and material cost. Nano-meter/radian resolution, large amplification ratio, repeatability, and stability are ensured due to the characteristics of the proposed monolithic micromanipulator. To investigate the motion range and decouple the micromanipulator's motions, the inverse kinematics is obtained using FEA. The performances of the developed micromanipulator are investigated in the real-time experiments via three feedback control methodologies, i.e. Proportional-Integral-Derivative control (PID), sliding mode control (SMC), and nonlinear disturbance observer-based sliding mode control (SMC-NDO). The role of the developed NDO is to compensate for the uncertain disturbances in the real-time experiments to achieve high precision manipulation tasks. In the end, the experimental results, including frequency, resolution, and several complex trajectory motion tracking analyses are presented, and precise manipulations can be guaranteed by the developed monolithic micromanipulator and control methodologies.

\vspace{-2mm}

\section{Mechanical Design}
As presented in Figure \ref{FIG_1}, the structure of the micromanipulator consists of a fixed base platform, six leaf-flexure-based parallelogram mechanisms, three flexure-based Scott-Russell amplification mechanisms, three flexure-based spherical joint modules, and a moving platform. Two sets of leaf parallelograms are incorporated into the input and output points of the Scott-Russell amplification mechanism as prismatic joints. This linearizes the motion of the Scott-Russell mechanism and increases the micromanipulator's stiffness. Because PEAs produce a very small displacement as a proportion of their length, mechanical displacement amplification is inevitably required for large displacement applications. The Scott-Russell amplification mechanism is a well established mechanical amplifier \cite{Ghafarian2020,Qin2013}. Additionally, it has the advantage of transforming into a horizontal input to a vertical output which is ideal for the compactness of the designed $\mathrm{Z\theta_{x}\theta_{y}}$ micromanipulator. The platform needs to generate rotational motions along two different axes. A spherical joint, which is capable of rotating in three different axes is adopted for the connection between the platform and the manipulation system. In comparison with previous designs, here the PEAs are placed outside of the micromanipulator not inside. Thus, different sizes of PEAs can be used for the proposed monolithic design without affecting the geometry and overall dimensions of the structural design. The overall dimensions of the proposed design are 201mm, 180mm, 75mm. Developments in 3D printing enable complex geometries to be easily and economically manufactured in multiple materials. Therefore, the proposed design is fabricated using high density and high accuracy 3D-printing from Acrylonitrile Butadiene Styrene (ABS). The mechanical and physical properties of ABS are presented in Table \ref{table_1}.

\vspace{-2mm}

\section{Workspace analysis}
The usable workspace is limited by material stresses. In order to maintain the stability, repeatability, and capabilities of the micromanipulator in precise manipulation, it is very important that the applied stress on the micromanipulator due to the load remains less than its tensile yield strength. Using the stress and safety factor analyses in FEA software (ANSYS), the maximum Von-Mises stress and the minimum safety factor occur when the input displacement of 90$\mu$m is applied to the micromanipulator as the input of the three PEAs simultaneously. The values of maximum Von-Mises stress and minimum safety factor corresponding to the 90$\mu$m inputs were 17.585MPa and 1.71, respectively. It is worth noting that the obtained minimum safety factor was 1.71, as the safety factor must always be greater than 1. Therefore, more input force/displacement could have been applied to obtain a larger workspace. However, the value of 1.71 is considered as a lower limit for the safety factor of the micromanipulator to avoid some deficiencies including creep, fatigue, and mechanical failure of the micromanipulator in the experiment.\\
\indent Three PEAs (Physik Instrumente P-843.60) with a maximum displacement of 90$\mu$m were assumed to be used for analysing the workspace based on the FEA. The inverse kinematics of the micromanipulator was computed by evaluating the motion of the system across 80 inputs, spanning the full input range, and performing a regression on the resulting outputs. The best fit inverse kinematics is computed as:

\begin{equation}
\textbf{$J^{-1}$}=\resizebox{0.5\hsize}{!}{$\begin{pmatrix}
-0.1909	& 0.0001	& 0.0110\\
-0.1877	& -0.0095	& -0.0053\\
-0.1875 & 0.0093 & -0.0056\\
	\end{pmatrix}$}
\end{equation}

\begin{figure}[!b]\centering
	\begin{tabular}{c}
		\hbox{\hspace{-0.5em}\includegraphics[width=9cm]{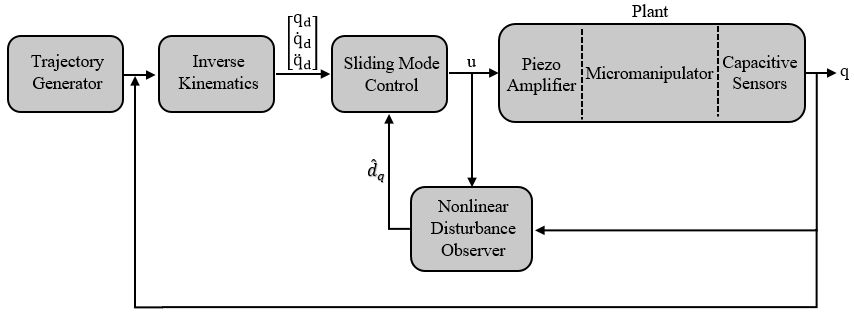}} \\ [0ex]
	\end{tabular}
	\caption{Block diagram of the proposed SMC-NDO control methodology}\label{FIG_3}
\end{figure}

The resultant workspace of computational analysis is presented in Figure \ref{FIG_2}. The values of motions along $\mathrm{Z}$, $\mathrm{\theta_x}$, and $\mathrm{\theta_y}$ axes are $\pm$238.5$\mu$m, $\pm$4830.5$\mu$rad, and $\pm$5486.2$\mu$rad, respectively. Since the determinant of the Jacobian matrix is non-zero, therefore there is no singularity in the workspace of the micromanipulator.

\begin{figure}[!b]\centering
	\begin{tabular}{c}
		\hbox{\hspace{-0.5em}\includegraphics[width=8.75cm]{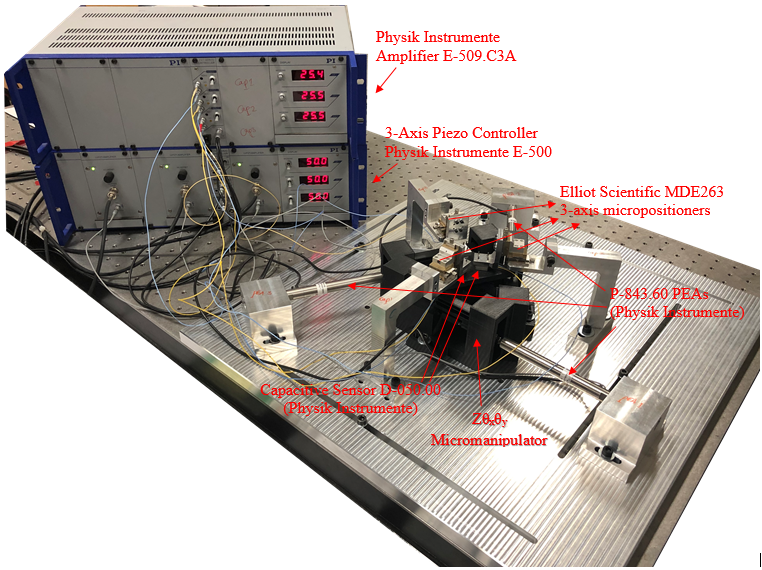}} \\ [0ex]
		\hbox{\hspace{-0.5em}\includegraphics[width=8.75cm]{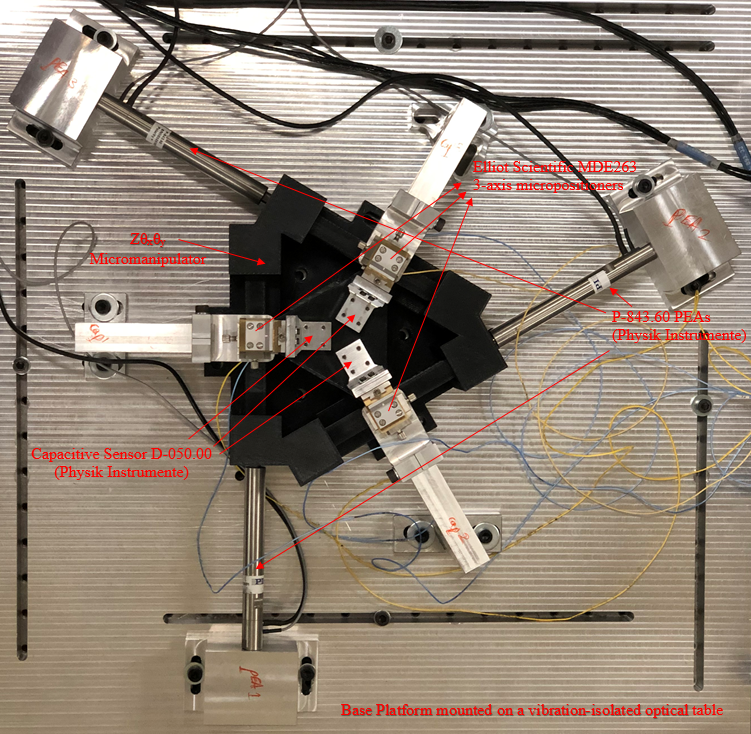}} \\ [0ex]	
	\end{tabular}
	\caption{Schematic diagram of sensing, control, and experimental research facility}\label{FIG_4}
\end{figure}

\begin{figure}[!t]\centering
	\begin{tabular}{c}
		\hbox{\hspace{0em}\includegraphics[width=6.5cm]{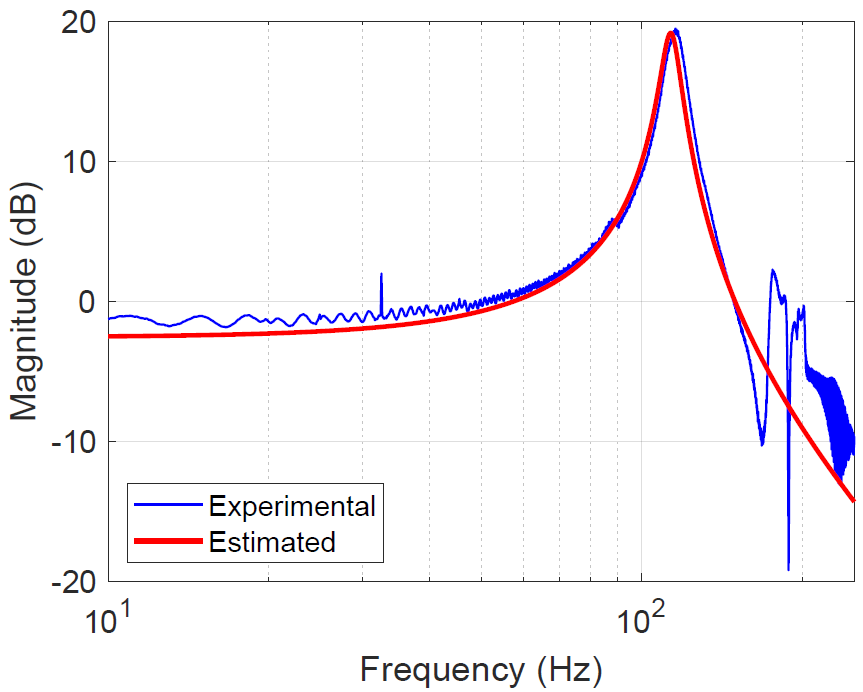}} \\ [0ex]
		\hbox{\hspace{2em}a} \\ [0ex]
		\hbox{\hspace{0em}\includegraphics[width=6.5cm]{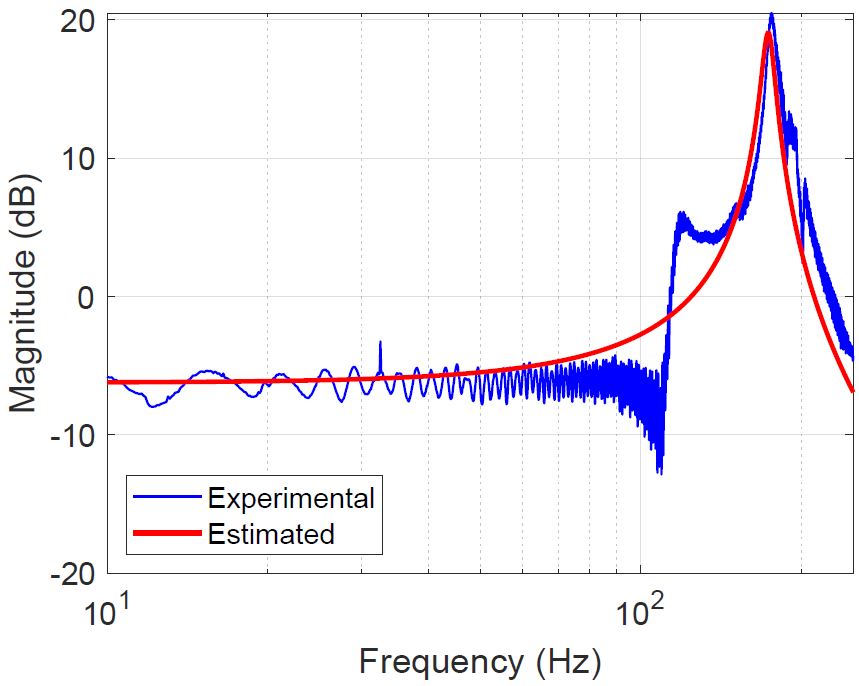}} \\ [0ex]
		\hbox{\hspace{2em}b} \\ [0ex]
		\hbox{\hspace{0em}\includegraphics[width=6.5cm]{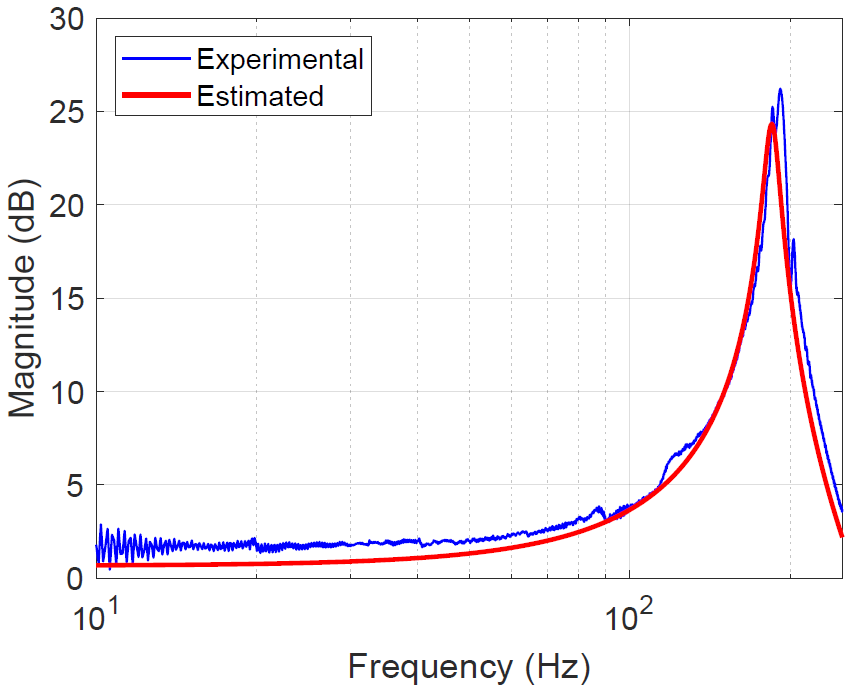}} \\ [0ex]
		\hbox{\hspace{2em}c} \\ [0ex]
	\end{tabular}
	\caption{Modal analysis of the micromanipulator: (a) $\mathrm{Z}$-resonant frequency (b) $\mathrm{\theta_x}$-resonant frequency (c) $\mathrm{\theta_y}$-resonant frequency}\label{FIG_5}
	\vspace{-1mm}
\end{figure}

\begin{figure*}[!t]\centering
	\begin{tabular}{ccc}
		\hbox{\hspace{-3em}\includegraphics[width=6.5cm]{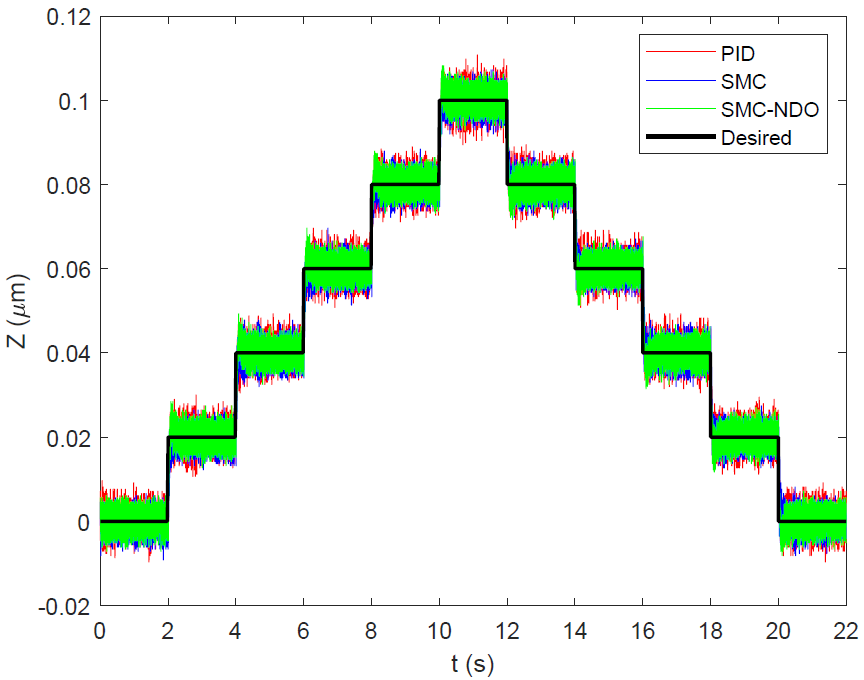}} &
		\hbox{\hspace{-1em}\includegraphics[width=6.5cm]{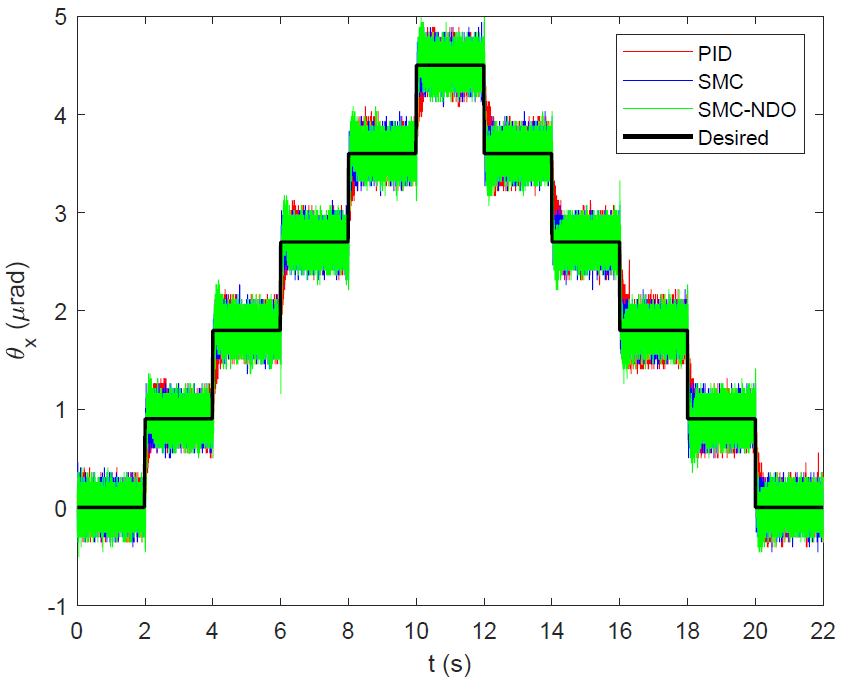}}  &
		\hbox{\hspace{-1em}\includegraphics[width=6.5cm]{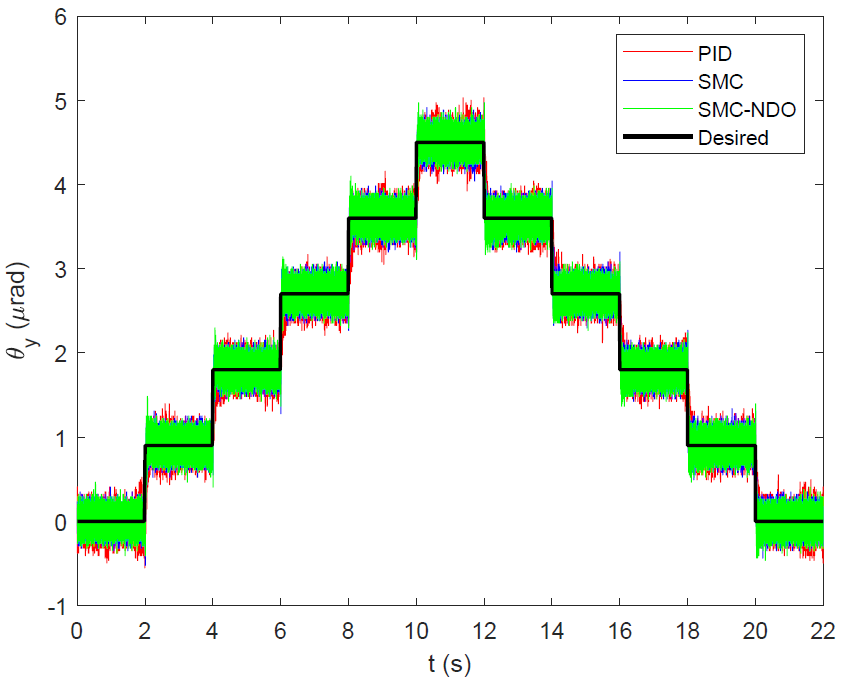}}  \\ 	[0ex]
	\end{tabular}
	\caption{System response to the applied staircase input}\label{FIG_res}
	\vspace{-5mm}
\end{figure*}

\vspace{-2mm}
\section{Controller design}
The accuracy of a manipulation task is very dependent on eliminating undesired disturbances affecting the system. Therefore, in this section, establishment of control methodologies will be presented for the utilization in the experimental study to overcome this challenging issue for precise micro/nano manipulation applications.

\vspace{-2mm}
\subsection{Sliding Mode Control (SMC)}
In general, flexure-based micromanipulators are structures comprising solid links and flexure hinges. The monolithic construction reduces assembly errors and guarantees high accuracy. A lumped parameter dynamic model for a flexure-based micromanipulator can be given by:

\vspace{-2.5mm}
\begin{align}
m_{q}\ddot{q}+c_q\dot{q}+k_qq=d_q+u_q
\end{align}

\noindent where $q$ represents the three output motions of the micromanipulator (Z, $\mathrm{\theta_x}$, $\mathrm{\theta_y}$), and $m_{q}$, $c_{q}$, and $k_{q}$ are the equivalent mass, damping and stiffness in the corresponding axis. $d_q$ and $u_q$ are the lumped disturbance and control action in a given axis, respectively. A state-space representation of Eq. (2) can be obtained by considering $x_{1}=q$ and $x_{2}=\dot{q}$. The tracking error between the actual and desired motions can be introduced as follows:

\vspace{-4mm}
\begin{align}
e=x_{1}-{{x}_{1}}_{des}
\end{align}
\vspace{-6mm}

There are two steps in the design of an SMC. The first step is designing a sliding surface so that the plant restricted to the sliding surface has a desired system response. This means the state variables of the plant dynamics are constrained to satisfy another set of equations which define the so-called switching surface. The second step is constructing a switched feedback gains necessary to drive the plant's state trajectory to the sliding surface. These strategies are built on the generalized Lyapunov stability theory. In this study, a PID sliding surface was selected to reduce the steady-state error and improve the transient response speed. The selected sliding surface is given by:

\vspace{-4mm}
\begin{align}
s=\lambda_{p}e+\lambda_{i}\int_{0}^{t} e dt+\lambda_{d}\dot{e}
\end{align}

\noindent where $\lambda_{p}$, $\lambda_{i}$, and $\lambda_{d}$ are positive proportional, integral, and derivative parameters to be selected, respectively. The SMC law is intended to force the tracking error $e$ to approach the sliding surface and then proceed to the origin along the sliding surface. Hence, the sliding surface needs to be stable, and therefore yields to the fact that $\dot{s}=0$. Additionally, the control action of the SMC approach consists of two parts; equivalent control and switching control as follows:

\vspace{-4mm}
\begin{align}
u=u_{eq}+u_{sw}
\end{align}

Thus, it can be realized that to be able to force the actual motions to converge to the desired ones, the equivalent control and switching control need to be as follows:

\begin{equation}
\begin{tabular}{c}
$u_{eq}=c{x}_2+kx_{1}+m\ddot{x}_{1_{des}} - m \frac{\lambda_{i}}{\lambda_{d}} e - m \frac{\lambda_{p}}{\lambda_{d}} \dot{e} - \hat{d}$ \\
$u_{sw}=-a_{1}s-a_{2}sgn(s)$
\end{tabular}
\end{equation}

\noindent where $a_{1}$ and $a_{2}$ are positive constants. The term $\hat{d}$ is the estimation of the disturbance which will be derived later. The deviation between the actual and estimated disturbance is defined by, $\tilde{d}$, and can be given as follows:

\begin{equation}
\tilde{d}=d-\hat{d}
\end{equation}

% star
\begin{figure}[!b]\centering
	\begin{tabular}{ccc}
		\hbox{\hspace{0em}\includegraphics[width=8.5cm]{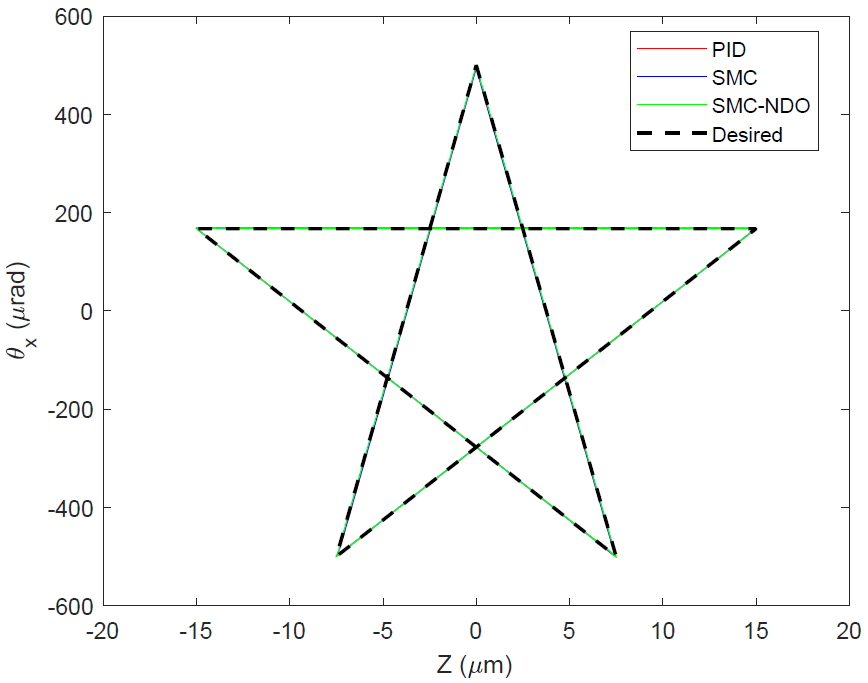}} \\ 	[0ex]
	\end{tabular}
	\caption{Star trajectory tracking results}\label{FIG_star}
\end{figure}

\begin{figure*}[!t]\centering
	\begin{tabular}{ccc}
		\hbox{\hspace{-3em}\includegraphics[width=6.5cm]{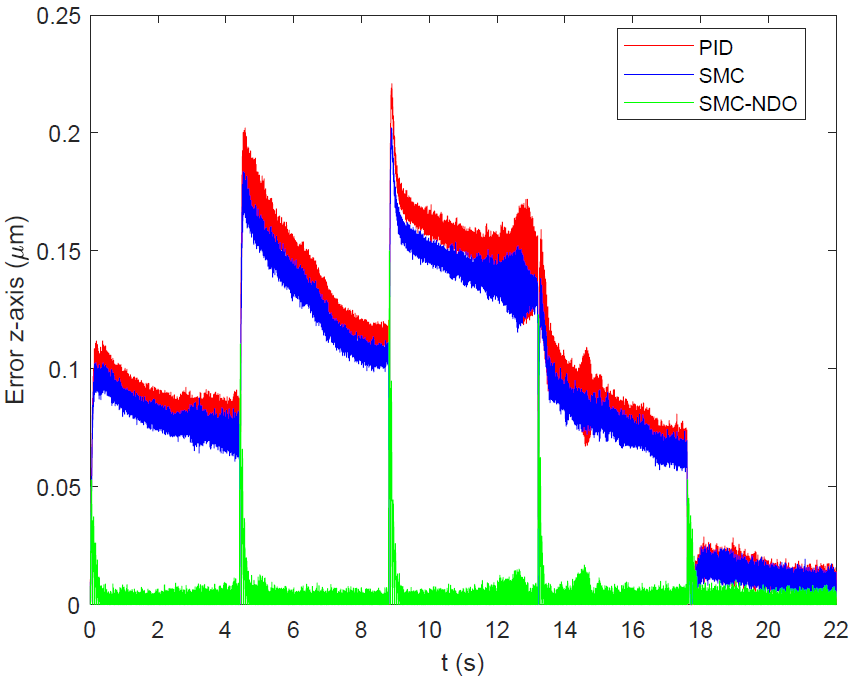}} &
		\hbox{\hspace{-1em}\includegraphics[width=6.5cm]{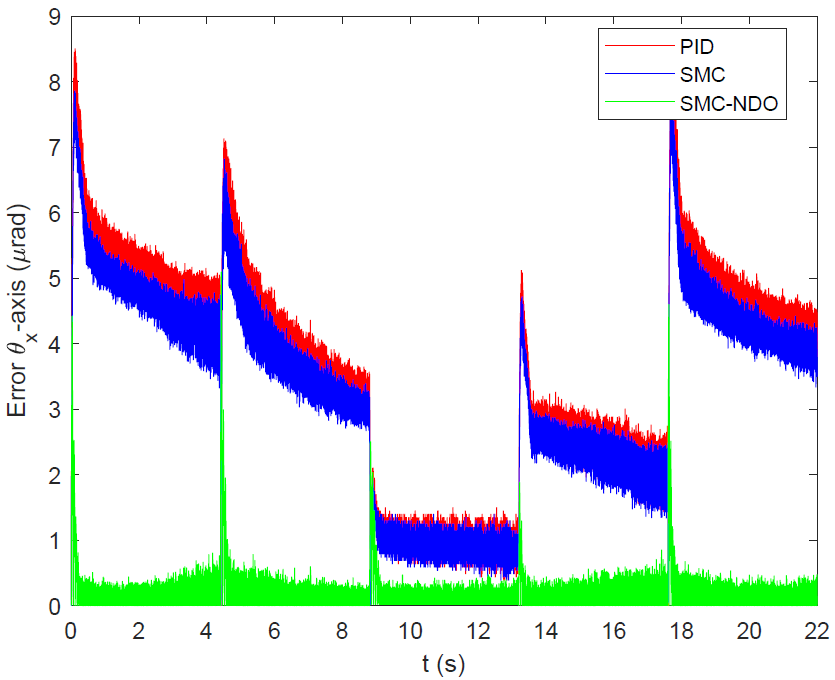}}  &
		\hbox{\hspace{-1em}\includegraphics[width=6.5cm]{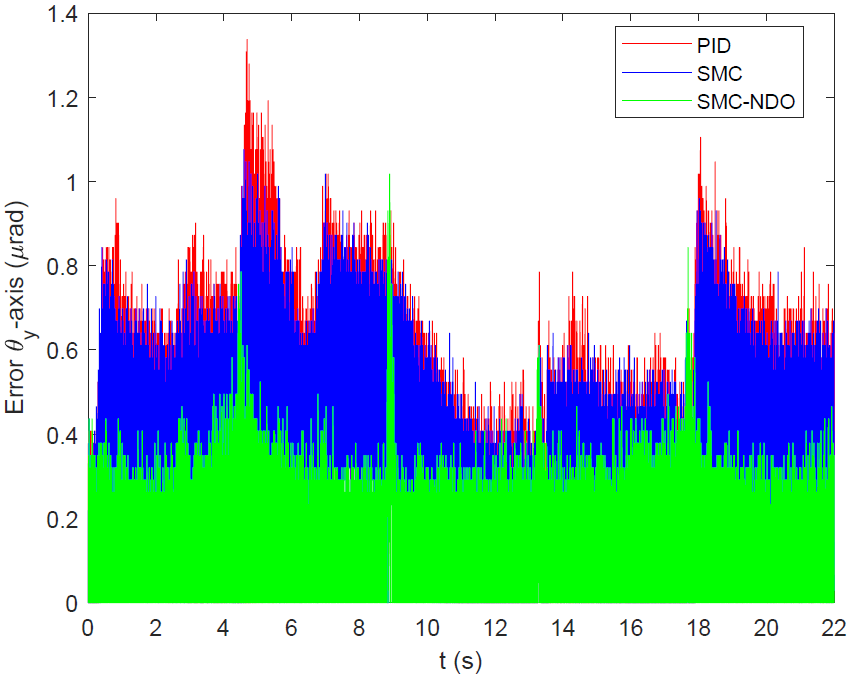}}  \\ 	[0ex]
	\end{tabular}
	\caption{Experimental control tracking errors in the star trajectory}\label{FIG_star_error}
	\vspace{-5mm}
\end{figure*}

Lyapunov stability function is used to verify the design performance of the SMC. In order to have a stable controller, it is required that $V$ must be bounded, which means that it is important to prove $\dot{V} \leq 0$. Therefore, the derivative of Lyapunov function is derived as follows:

\vspace{-4mm}
\begin{align}
\dot{V}=s\dot{s}=s\{\frac{\lambda_{d}}{m}u_{sw}+\frac{\lambda_{d}}{m}d-\frac{\lambda_{d}}{m}\hat{d}\}=s\{-k_1s \notag\\ -k_2sgn(s)+\frac{\lambda_{d}}{m}\tilde{d}\}=-k_1s^2-k_2|s|\notag\\ +\frac{\lambda_{d}}{m}\tilde{d}s \leq -k_1s^2-|s|(k_2-\frac{\lambda_{d}}{m}\tilde{d})
\end{align}

\noindent where $k_1$ and $k_2$ are equal to $\frac{\lambda_{d}a_1}{m}$ and $\frac{\lambda_{d}a_2}{m}$, respectively. The stability of the designed controller is guaranteed, if the inequality $k_2 \geq \frac{\lambda_{d}}{m}\tilde{d}$ is established. Furthermore, the discontinuity of the signum function in the SMC law can cause chattering. Therefore, in order to avoid this problem, the discontinuous signum function is replaced by the continuous hyperbolic tangent function ($tanh$). In other words, the function is used as an approximator of the signum function.

\vspace{-3mm}
\subsection{Nonlinear Disturbance Observer (NDO)}
An NDO is designed to estimate the undesired disturbance $d_q$, introduced in the dynamic model, Eq. (2). $d_q$ represents all disturbances that may be caused by unmodeled dynamics. The objective is to design an observer such that the estimation $\hat{d}$ yielded by the observer exponentially approaches the disturbance $d$ under any $q$ and $\dot{q}$.\\
The dynamic model of the micromanipulator, Eq. (2), can be rewritten in the general description of a form as follows:

\begin{equation}
\dot{\textbf{x}}=f(\textbf{x})+g_1(\textbf{x})u+g_2(\textbf{x}){d}
\end{equation}

\noindent where $\textbf{x}$ is a representation of the state variables $x_1$ and $x_2$. Besides, functions $f$, $g_1$, and $g_2$ can also be found to be as follows:

\vspace{-2mm}
\begin{equation}
\begin{tabular}{c}
$g_1(x)=g_2(x)=\frac{1}{m_q}$ \\
$f(x)=-\frac{c_q}{m_q}x_2-\frac{k_q}{m_q}x_1$
\end{tabular}
\end{equation}

The NDO is designed according to the following formulations \cite{Al-Jodah2020a,Al-Jodah2020b}:

\vspace{-2mm}
\begin{equation}
\begin{tabular}{c}
$\hat{d}=z+p(\textbf{x})$ \\
$\dot{z}=-l(\textbf{x})g_2(\textbf{x})z-l(\textbf{x})\{g_2(\textbf{x})p(\textbf{x})+f(\textbf{x})+g_1(\textbf{x})u\}$
\end{tabular}
\end{equation}

\noindent where $\hat{d}$ and $z$ are the estimate of the undesired disturbance and the internal state of the nonlinear observer, respectively, and $p(\textbf{x})$ is a nonlinear function to be designed. The nonlinear observer gain $l(\textbf{x})$ is defined as follows:

\begin{equation}
l(\textbf{x})=\frac{\partial p(\textbf{x})}{\partial \textbf{x}}
\end{equation}

\noindent It can be shown that $\hat{d}$ approaches $d$ exponentially if $p(\textbf{x})$ is chosen such that:

\begin{equation}
\dot{\tilde{d}}+\frac{\partial p(\textbf{x})}{\partial \textbf{x}}g_2(x)\tilde{d}=0
\end{equation}

\noindent Therefore, if $p(\textbf{x})$ is chosen as $lx_2$, it can be realized from Eq. (13) that the designed NDO yields to following form in the time domain: 

\begin{equation}
\tilde{d}(t)=\tilde{d}(0)exp(-\frac{l}{m}t)
\end{equation}

\noindent According to Eq. (14), if $l$ is chosen to be a positive constant, when time goes to infinity the disturbance estimation error, $\tilde{d}$, converges to zero. Therefore, the designed NDO is globally exponentially stable.

% logo
\begin{figure}[!b]\centering
	\begin{tabular}{ccc}
		\hbox{\hspace{0em}\includegraphics[width=8.5cm]{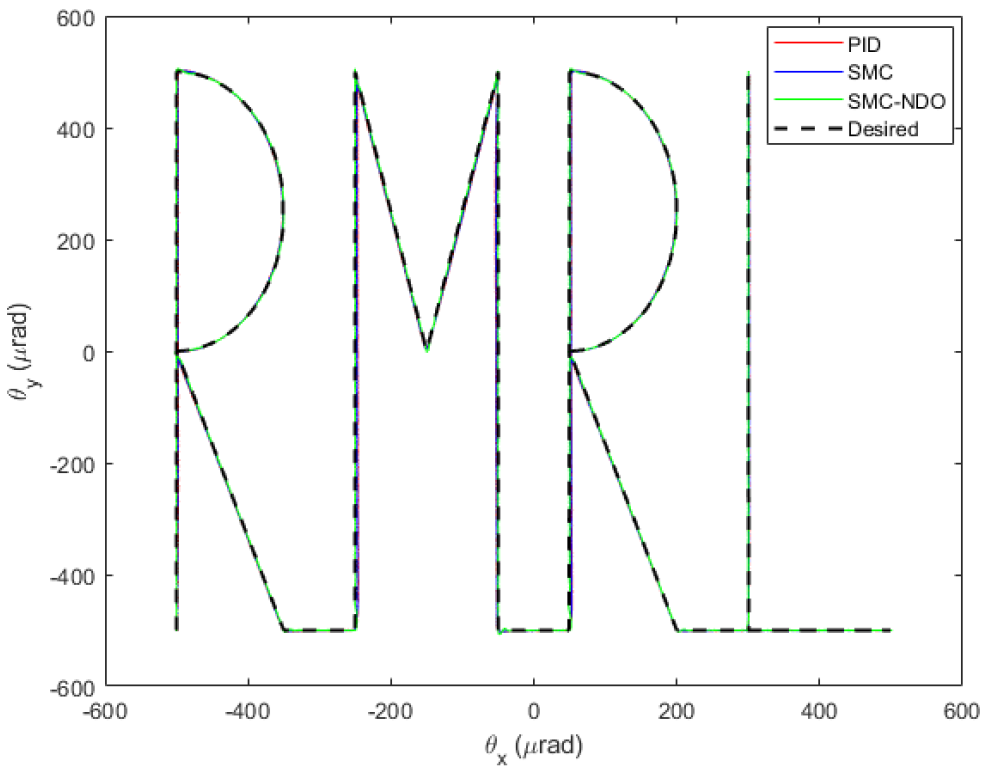}} \\ 	[0ex]
	\end{tabular}
	\caption{RMRL logo trajectory tracking results}\label{FIG_logo}
	\vspace{0mm}
\end{figure}

\begin{figure*}[!t]\centering
	\begin{tabular}{ccc}
		\hbox{\hspace{-3em}\includegraphics[width=6.5cm]{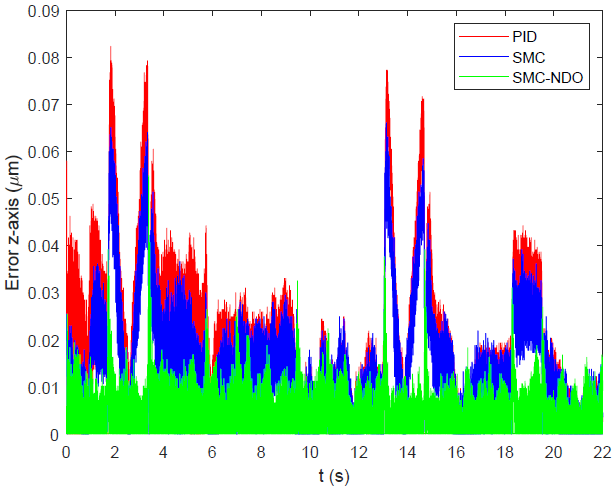}} &
		\hbox{\hspace{-1em}\includegraphics[width=6.5cm]{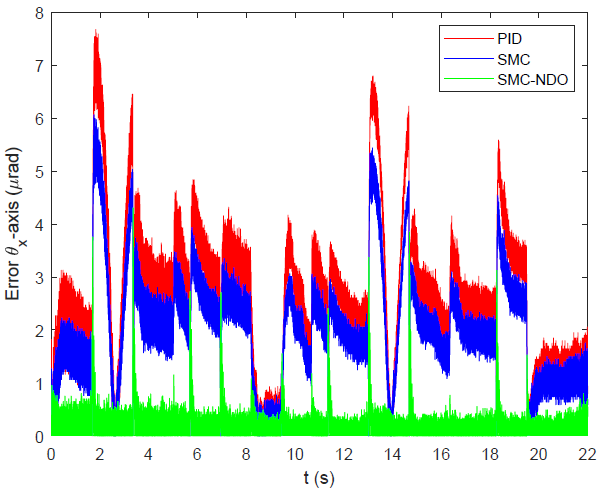}}  &
		\hbox{\hspace{-1em}\includegraphics[width=6.5cm]{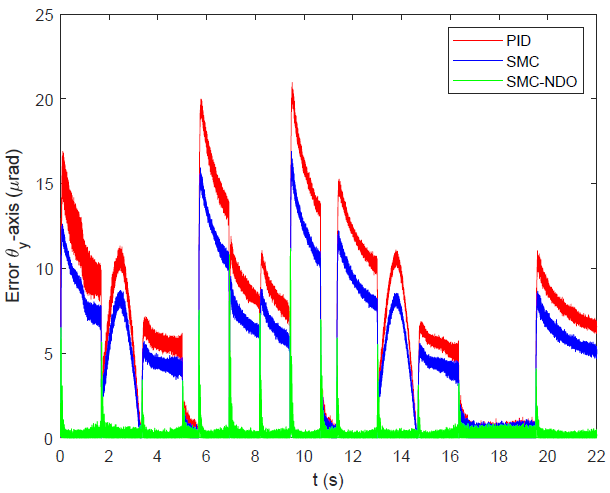}}  \\ 	[0ex]
	\end{tabular}
	\caption{Experimental control tracking errors in the RMRL logo trajectory}\label{FIG_logo_error}
	\vspace{-3mm}
\end{figure*}

\vspace{-2mm}
\subsection{Robustness Analysis of SMC-NDO}
The robustness of the proposed SMC-NDO is proven in this section. The feedback control of the system described by Eq. (2) is robust, if control law (5) is applied and it satisfies the conditions of:

\vspace{-2mm}

\begin{equation}
\begin{tabular}{ccc}
$\lim\limits_{s\to0^{+}} \dot{s} < 0$ & $,$ & $\lim\limits_{s\to0^{-}} \dot{s} > 0$
\end{tabular}
\end{equation}

\vspace{0mm}

\noindent The above-mentioned conditions illustrate that if the phase trajectory around $s=0$ points to the $s$ hyper-plane, then it will enter the sliding mode when the hyper-plane $s=0$.\\
\indent By substituting control law (5) into Eq. (2), the equation of motion of the closed-loop system is obtained as:

\vspace{-4mm}

\begin{align}
\ddot{q} & = \frac{1}{m_{q}} \Big\{\tilde{d}_{q}-a_{1}s-a_{2}tanh(s) \Big\}+ \Big\{\ddot{q}_{des}-\frac{\lambda_{i}}{\lambda_{d}}e-\frac{\lambda_{p}}{\lambda_{d}}\dot{e} \Big\}
\end{align}

\vspace{-2mm}

\noindent Moreover, by substituting Eq. (16) into the differentiation of Eq. (4), the dynamic sliding mode equation is obtained as:

\vspace{-4mm}

\begin{align}
\dot{s} & = \lambda_{p}\dot{e}+\lambda_{i}e+\lambda_{d}\ddot{e} \notag\\& =\lambda_{p}\dot{e}+\lambda_{i}e+\lambda_{d}(\ddot{q}-\ddot{q}_{des}) \notag\\& \approx \tilde{d}_{q}-a_{1}s-a_{2}tanh(s)
\end{align}

\vspace{-1mm}

\noindent Utilizing the two conditions (15), and the stability condition that was obtained in Section III.A, ($a_2 \geq \tilde{d}$), it is trivial that the robustness conditions are satifisfied as following:

\vspace{-1mm}

\begin{equation}
	\begin{aligned}
\lim\limits_{s\to0^{+}} \dot{s} = \tilde{d}_{q}-a_{2}tanh(s) < 0 \\
\lim\limits_{s\to0^{-}} \dot{s} = \tilde{d}_{q}-a_{2}tanh(s) > 0
\end{aligned}
\end{equation}

\vspace{0mm}

\noindent Thus, the phase trajectory around $s=0$ enters the sliding mode in the hyper-plane and $s=0$, so the PID sliding surface (4) is:

\vspace{-5mm}

\begin{align}
s = (\lambda_{p}+\lambda_{i}s^{-1}+\lambda_{d}s)e=0
\end{align}

\vspace{0mm}

\noindent Using Eq. (19) and having the knowledge of $\dot{e}=\dot{x}_{2}-\dot{x}_{2_{des}}$, Eq. (9) is rewritten as following:

\vspace{-1mm}

\begin{equation}
\begin{tabular}{c}
$x_1=x_{1_{des}}+e$ \\
$x_2=x_{2_{des}}-\frac{\lambda_{p}}{\lambda_{d}}e-\frac{\lambda_{i}}{\lambda_{d}}s^{-1}e$
\end{tabular}
\end{equation}

\vspace{0mm}

\noindent Thus, Eq. (20) becomes the state equation when the system enters the sliding mode. We know that the second-order system Eq. (2) can be expressed by the first-order state Eq. (20) and that the dynamic characteristics of the system are independent of $d(t)$, therefore, SMC-NDO is highly robust when the system enters the sliding mode, thereby ensuring that the system exhibits a good dynamic response and stability.

\vspace{0mm}

\begin{figure}[!b]\centering
	\begin{tabular}{ccc}
		\hbox{\hspace{0em}\includegraphics[width=8.5cm]{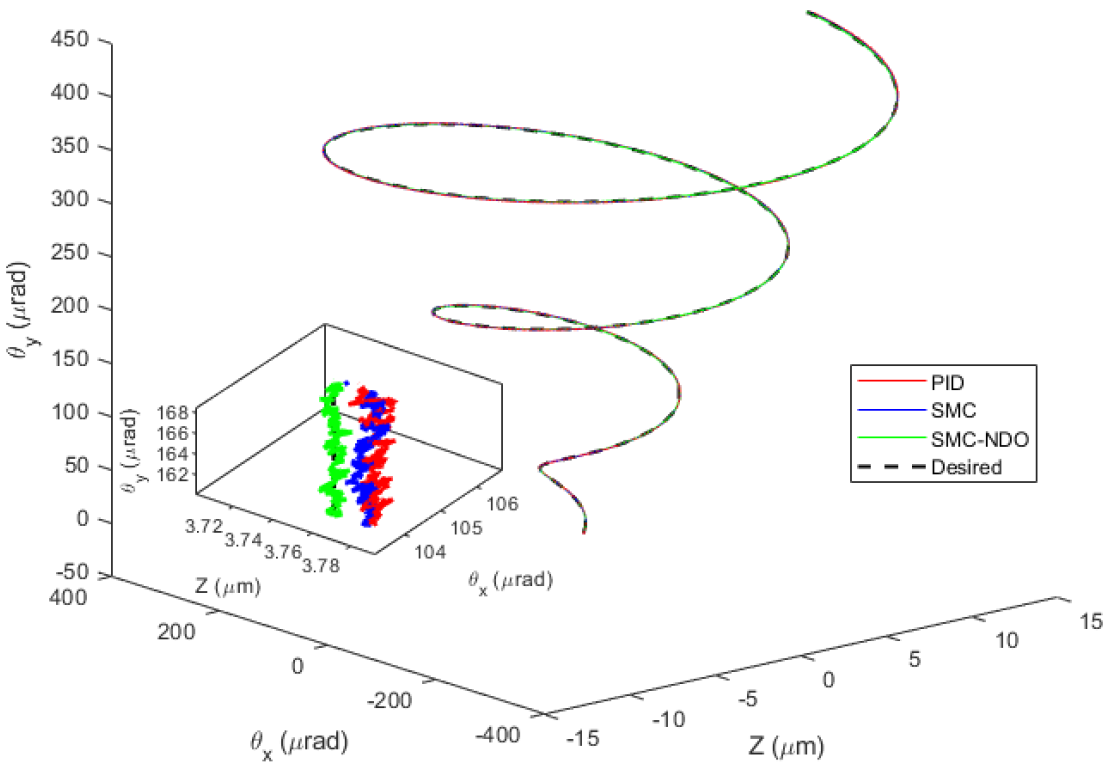}} \\ 	[0ex]
	\end{tabular}
	\caption{3D-Archimedean spiral trajectory tracking results}\label{FIG_tornado}
\end{figure}

\vspace{-1mm}

\subsection{Time Convergence Stability Analysis of SMC-NDO}
The time convergence of the sliding mode equation is investigated in this section when the disturbance observer and chattering reduction methods are used.

\textbf{Lemma.} \cite{Mobayen2015} Assume that a continuous definite function $V(t)\geq0$ satisfies the subsequent differential inequality $\dot{V}(t)+\beta_{1}V^\alpha (t)+\beta_{2}V(t) \leq 0$, where $\beta_{1}$, $\beta_{2}$, and $\alpha$ are the positive constants, and the range of $\alpha$ is between $0$ to $1$. Based on which the function $V(t)$ converges to zero in finite time given as:

\vspace{-2mm}

\begin{equation}
t_s=\frac{1}{\beta_{2}(1-\alpha)}\ln\frac{\beta_{2}V^{(1-\alpha)}(0)+\beta_1}{\beta_1}
\end{equation}

\textbf{Theorem.} Consider system described by Eq. (2). If the surface manifold and control law are chosen according to Eqs. (4) and (6), then sliding mode enforcement along the sliding surface, Eq. (4), can be validated. As a result, the system states will converge to the desired reference trajectories and tracking errors will converge to zero in finite time.

\textbf{Proof of Theorem.} Consider the Lyapunov candidate functions of the form
as:

\vspace{-3mm}
\begin{equation}
V=\frac{1}{2}s^2
\end{equation}

\noindent Taking the time derivative of Eq. (22) and utilizing Eq. (17) yield to:

\vspace{-4mm}
\begin{equation}
\dot{V}=s\{\tilde{d}-a_{1}s-a_{2}tanh(s)\} \leq -a_{1}s^2-a_{2}|s|
\end{equation}

\noindent Let $\beta_{1}=\sqrt{2}a_2$, $\beta_{2}=2a_1$, and $\alpha=\frac{1}{2}$, then Eq. (23) can be rewritten as:

\vspace{-4mm}
\begin{equation}
\dot{V}(t)+\beta_{1}V^\alpha (t)+\beta_{2}V(t) \leq 0
\end{equation}

\noindent Then, according to Lemma, the differential inequality (24) yields the finite-settling time as represented by Eq. (21), and that completes the proof.\\
Finally, Figure \ref{FIG_3} provides the information on how the SMC-NDO was implemented in the real-time experiment to achieve remarkable performances from the micromanipulator.

\begin{table}[!b]
	\renewcommand{\arraystretch}{1.3}
	\caption{Control parameters}
	\centering
	\label{table_2}
	\resizebox{0.85\columnwidth}{!}{
	\begin{tabular}{l l l}
		\hline\hline \\[-3mm]
		\textbf{Description} & \textbf{Parameter} & \textbf{Value} \\[0ex] \hline 
		\multicolumn{3}{c}{\textbf{PID}} \\[0ex] \hline 
		\multirow{3}{*}{PID constants} & $k_p$ & 50 \\[0ex] 
		{} & $k_i$ & 17.5e6 \\[0ex] 
		{} & $k_d$ & 0 \\[0ex] \hline 
		\multicolumn{3}{c}{\textbf{SMC}} \\[0ex] \hline 
		\multirow{3}{*}{PID sliding surface constants} & $\lambda_p$ & 50 \\[0ex] 
		{} & $\lambda_i$ & 13e6 \\[0ex] 
		{} & $\lambda_d$ & 100 \\[0ex] \cline{2-3} 
		\multirow{2}{*}{Switching control constants} & $a_1$ & 2.5 \\[0ex] 
		{} & $a_2$ & 0.6 \\[0ex] \hline  	
		\multicolumn{3}{c}{\textbf{SMC-NDO}} \\[0ex] \hline 
		Nonlinear observer gain & $l$ & 100 \\[0ex] \hline \hline  
	\end{tabular}
	}
	\vspace{-4mm}
\end{table}

% tornado

\begin{figure*}[!t]\centering
	\begin{tabular}{ccc}
		\hbox{\hspace{-3em}\includegraphics[width=6.5cm]{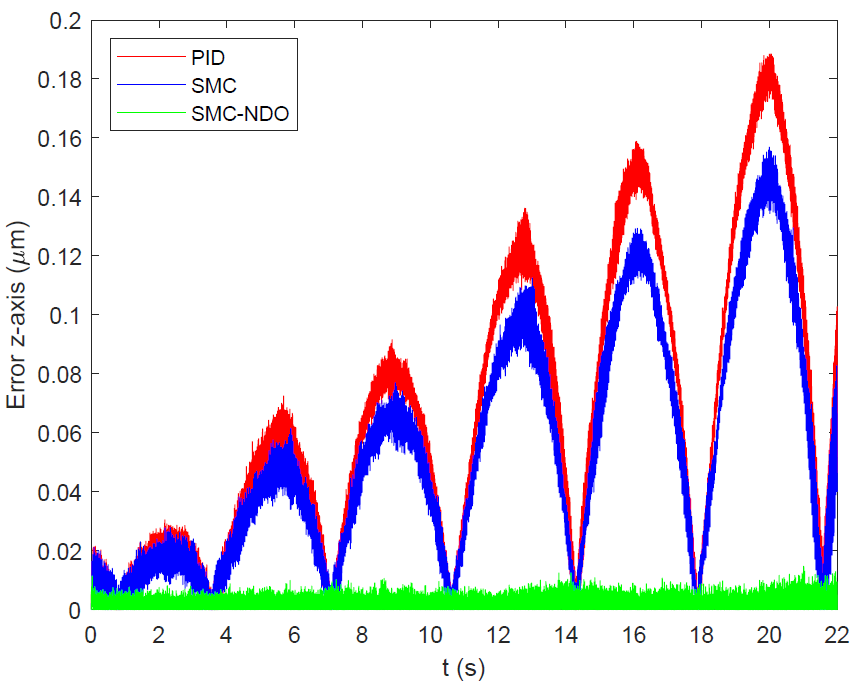}} &
		\hbox{\hspace{-1em}\includegraphics[width=6.5cm]{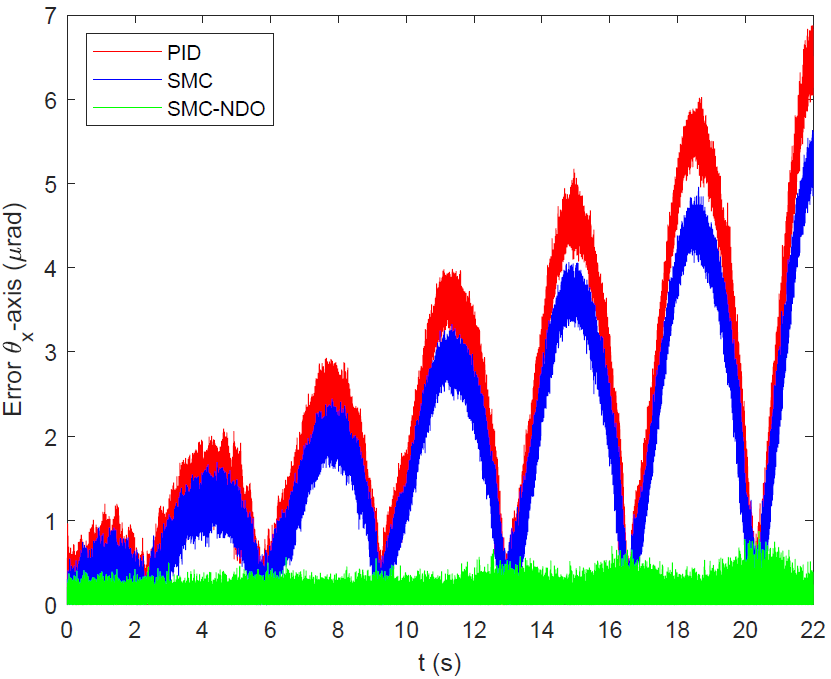}}  &
		\hbox{\hspace{-1em}\includegraphics[width=6.5cm]{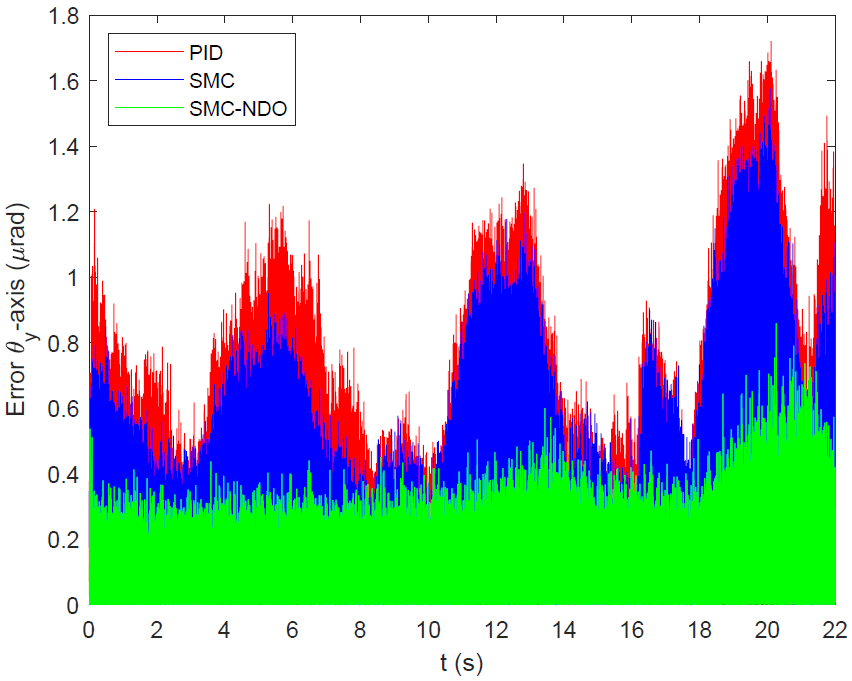}}  \\ 	[0ex]
	\end{tabular}
	\caption{Experimental control tracking errors in the 3D-Archimedean spiral trajectory}\label{FIG_tornado_error}
	\vspace{-4mm}
\end{figure*}

\vspace{-2mm}
\section{Experiments and Results}
Experimental tests were carried out to investigate the mechanical and dynamic performances of the developed flexure-based micromanipulator, as well as the efficiency of the proposed control methodologies. The schematic diagram of the experimental set-up is presented in Figure \ref{FIG_4}. The basic operation of the flexure-based micromanipulator was to change the drive voltage from the voltage control unit (amplifier module from Physik Instrumente) to the three PEAs from Physik Instrumente (model P-843.60) to further drive the micromanipulator. During installation of the PEAs, pre-compression forces were applied to keep the actuators' tip and the micromanipulator in contact. These PEAs are multilayer PZT stacked ceramic translators capable of 90$\mu$m displacement corresponding to a range of operating voltage from 0 to 100$\mathrm{V}$. The position of the moving platform was measured by three capacitive sensors (Physik Instrumente D-050.00). The capacitive sensors had a circular active area with a 4 mm radius surrounded by an annular guard ring, with a specified working range of 50$\mu$m. The electronic interface (Physik Instrumente E-509.C3A) to the capacitive sensors  provided a distance measurement as an analog voltage, which was recorded at the control computer with a 16-bit analog-to-digital converter (ADC) and at the rate of $10\mathrm{kHz}$. In order to reduce the external vibration disturbances, all experimental tests were performed on a vibration-isolated optical table.

\begin{figure*}[!t]\centering
	\begin{tabular}{ccc}
		\hbox{\hspace{-2em}\includegraphics[height=5.2cm,width=6.2cm]{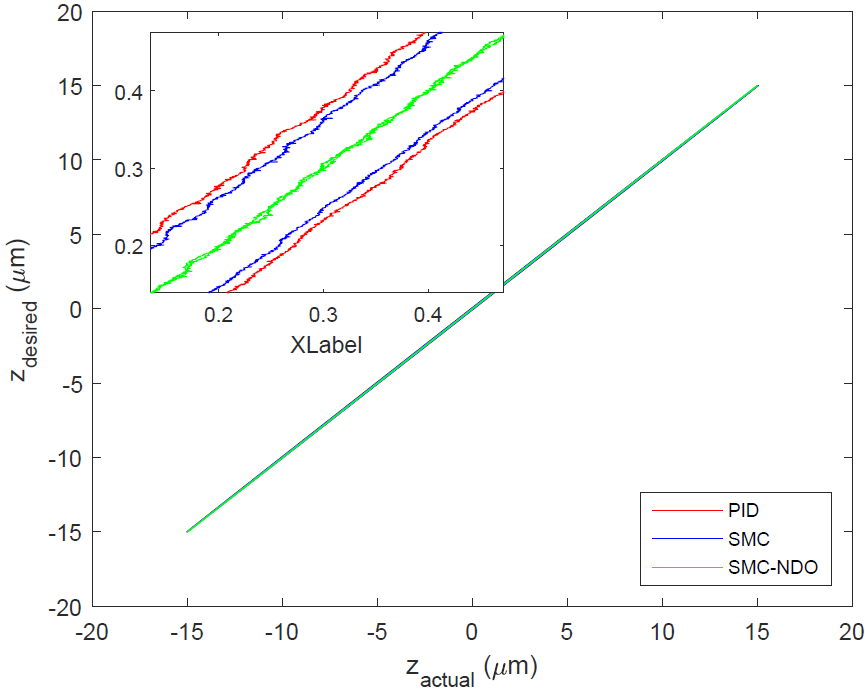}} &
		\hbox{\hspace{-1.1em}\includegraphics[height=5.25cm,width=6.25cm]{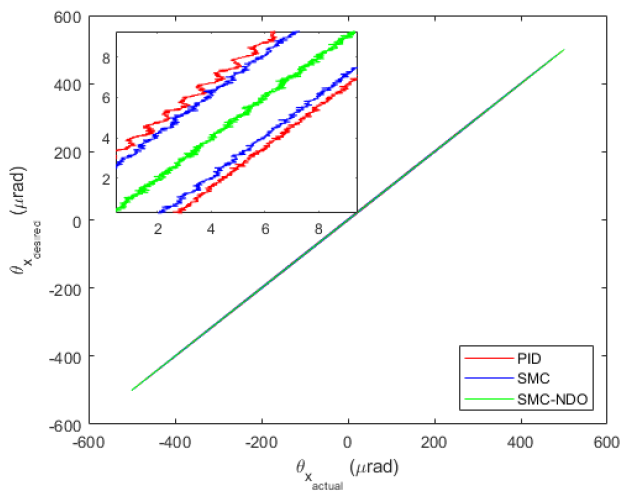}}  &
		\hbox{\hspace{-1em}\includegraphics[height=5.25cm,width=6.25cm]{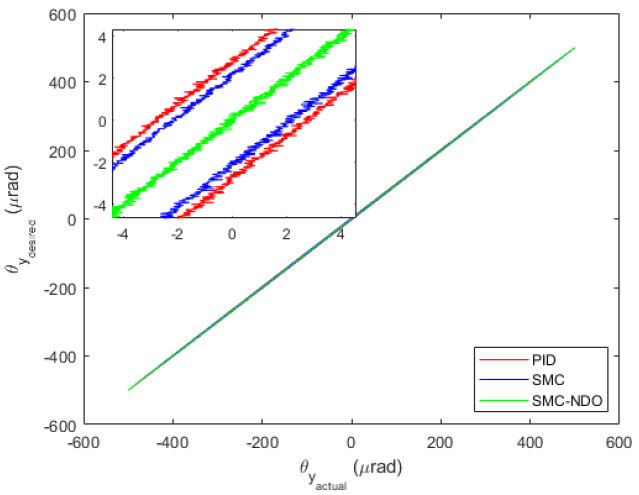}}  \\ 	[0ex]
	\end{tabular}
	\caption{Hysteresis cancellation of different control methodologies}\label{FIG_hyst}
\end{figure*}

\begin{table*}[htpb]
	\renewcommand{\arraystretch}{1.3}
	\caption{Summary of the numerical results of trajectory tracking and hysteresis compensation}
	\centering
	\label{table_3}
	\resizebox{\textwidth}{!}{%
	\begin{tabular}{cccccccc}
		\hline\hline \\[-3mm]
		\multirow{3}{*}{} & \multirow{3}{*}{\textbf{Controller}} & \multicolumn{2}{c}{\textbf{Z-axis}} & \multicolumn{2}{c}{\textbf{$\theta_{x}$-axis}} & \multicolumn{2}{c}{\textbf{$\theta_{y}$-axis}} \\[0ex] \cline{3-8}
		\multicolumn{2}{c}{} & \multicolumn{6}{c}{\textbf{Trajectory tracking}} \\[0ex] \cline{3-8}
		\multicolumn{2}{c}{} & \textbf{RMSE ($\mathrm{\mu m}$)} & \textbf{$\%$Improvement (SMC-NDO/PID)} & \textbf{RMSE ($\mathrm{\mu rad}$)} & \textbf{$\%$Improvement (SMC-NDO/PID)} & \textbf{RMSE ($\mathrm{\mu rad}$)} & \textbf{$\%$Improvement (SMC-NDO/PID)} \\[0ex] \hline 
		\multirow{3}{*}{\textbf{Star}} & \textbf{PID} & 0.1094 & \multirow{3}{*}{91.04} & 3.9298 & \multirow{3}{*}{90.80} & 0.3903 & \multirow{3}{*}{66.92} \\[0ex] 
		{} & \textbf{SMC} & 0.1012 & {} & 3.6336 & {} & 0.3596 & {} \\[0ex]  
		{} & \textbf{SMC-NDO} & 0.0098 & {} & 0.3615 & {} & 0.1291 & {} \\[0ex]  \hline
		
		\multirow{3}{*}{\textbf{RMRL logo}} & \textbf{PID} & 0.0239 & \multirow{3}{*}{74.06} & 2.9472 & \multirow{3}{*}{87.61} & 8.8958 & \multirow{3}{*}{91.07} \\[0ex] 
		{} & \textbf{SMC} & 0.0184 & {} & 2.2886 & {} & 6.9058 & {} \\[0ex]  
		{} & \textbf{SMC-NDO} & 0.0062 & {} & 0.3652 & {} & 0.7942 & {} \\[0ex]  \hline
		
		\multirow{3}{*}{\textbf{3D-Archimedean spiral}} & \textbf{PID} & 0.0819 & \multirow{3}{*}{96.83} & 2.7101 & \multirow{3}{*}{95.44} & 0.5296 & \multirow{3}{*}{76.25} \\[0ex] 
		{} & \textbf{SMC} & 0.0660 & {} & 2.1850 & {} & 0.4345 & {} \\[0ex]  
		{} & \textbf{SMC-NDO} & 0.0026 & {} & 0.1237 & {} & 0.1258 & {} \\[0ex] \hline  
		
		\multicolumn{2}{c}{} & \multicolumn{6}{c}{\textbf{Hysteresis compensation}} \\[0ex] \cline{3-8}
		\multicolumn{2}{c}{} & \textbf{$\%$Hysteresis max.} & \textbf{$\%$Improvement (SMC-NDO/PID $\&$ SMC)} & \textbf{$\%$Hysteresis max.} & \textbf{$\%$Improvement (SMC-NDO/PID $\&$ SMC)} & \textbf{$\%$Hysteresis max.} & \textbf{$\%$Improvement (SMC-NDO/PID $\&$ SMC)} \\[0ex] \hline
		
		\multirow{3}{*}{} & \textbf{PID} & 0.49 & \multirow{3}{*}{100} & 0.44 & \multirow{3}{*}{100} & 0.54 & \multirow{3}{*}{100} \\[0ex] 
		{} & \textbf{SMC} & 0.36 & {} & 0.38 & {} & 0.44 & {} \\[0ex]
		{} & \textbf{SMC-NDO} & 0 & {} & 0 & {} & 0 & {} \\[0ex]
		\hline \hline
	\end{tabular}
	} 
	\vspace{-4mm}
\end{table*}

\vspace{-2mm}
\subsection{System Identification}
System identification of the developed micromanipulator was conducted to investigate the dynamic characteristics. Hence, a sinusoidal sweep signal with the frequency linearly varying from $1\mathrm{Hz}$ to $250\mathrm{Hz}$ was applied to the PEAs input, one at a time, in order to excite the modes of vibration of the platform. The frequency responses to these inputs are presented in Figure \ref{FIG_5}. Resonance was observed in the response along the $\mathrm{Z}$, $\mathrm{\theta_x}$, and $\mathrm{\theta_y}$ axes occured at $113.4\mathrm{Hz}$, $173.1\mathrm{Hz}$, and $184.2\mathrm{Hz}$, respectively. Finally, three continuous-time transfer functions were identified from the frequency response along each motion axis and formulated as follows:

\begin{equation}
TF_z=\frac{3.774e05}{s^2 + 58.44 s + 5.08e05}
\end{equation}

\begin{equation}
TF_{\theta_x}=\frac{5.792e05}{  s^2 + 59.16 s + 1.186e06}
\end{equation}

\begin{equation}
TF_{\theta_y}=\frac{1.454e06}{  s^2 + 76.13 s + 1.346e06}
\end{equation}

Considering the transfer functions (15)-(17),  $m_q$, $c_q$, and $k_q$ can easily be calculated for implementing into to the SMC and SMC-NDO.\\
\indent In the next stage, i.e. closed-loop experiment, the control parameters were found by empirical rules based on observation of the response to reference and disturbance changes. These selected control parameters for the PID, SMC, and SMC-NDO control schemes are presented in Table \ref{table_2}.

\vspace{-2mm}
\subsection{Resolution characterization of the system}
To test the closed-loop system positioning resolution, the staircase response of the positioner was measured along each working axis. The magnitude of each step was chosen to be 20nm for the translational axis and 900nrad for the rotational axes. The obtained results from the three control techniques are presented in Figure \ref{FIG_res}. The best resolution result was captured utilizing the SMC-NDO control technique. The minimum resolution was found to be 8$\mathrm{nm}$, 500$\mathrm{nrad}$, and 460$\mathrm{nrad}$ along the $\mathrm{Z}$, $\mathrm{\theta_{x}}$, and $\mathrm{\theta_{y}}$ axes, respectively. The positioning resolution is limited by the measurement noise in motion axes, quantization in the measurement of rotation, the resolution of capacitive sensors, and the output resolution of the data acquisition card.

\vspace{-2mm}
\subsection{Motion tracking characterization of the system}
To further illustrate the motion abilities of the developed micromanipulator alongside with the designed control approaches, three complex curves tracking experiments were conducted. In which, a star, Robotics and Mechatronics Research Laboratory (RMRL) logo, and 3D-Archimedean spiral were selected and designed as the target tracking trajectories due to their high level of complexities. The tracking results are presented in Figures \ref{FIG_star} to \ref{FIG_tornado_error}.\\
The results of the trajectory tracking and tracking errors of the star trajectory implementing the three proposed control schemes are presented in Figures \ref{FIG_star} and \ref{FIG_star_error}. The best tracking error was obtained by implementing SMC-NDO and the maximum values of error were 14$\mathrm{nm}$ for the Z-axis, 700$\mathrm{nrad}$ and 494$\mathrm{nrad}$ for the $\mathrm{\theta_{x}}$ and $\mathrm{\theta_{y}}$ axes, respectively. Figure \ref{FIG_star} illustrates the projection of the tracked trajectory on the ZY-plane.\\
The RMRL logo was designed as a symbol of the most complex trajectory to verify the performance of the micromanipulator. The results of the trajectory tracking and tracking errors are presented in Figures \ref{FIG_logo} and \ref{FIG_logo_error}. The projection of the designed trajectory on the XY-plane is presented in Figure \ref{FIG_logo}. According to the results, the micromanipulator was capable of tracking the designed logo with high precision and accuracy. It must be noted that by using sensors with a higher resolution, more accurate motion tracking can be achieved even with the same control methodologies proposed in this work.\\
The tracking results of the 3D-Archimedean spiral are presented in Figures \ref{FIG_tornado} and \ref{FIG_tornado_error}. These two figures provide the difference between the desired and actual trajectories while implementing different control methodologies. Same as the previous trajectories tracking, the SMC-NDO control scheme was observed to outperform the SMC and PID controllers. Also, the SMC provided a better tracking performance than the PID control scheme. Considering SMC-NDO, the maximum tracking error along the Z, $\mathrm{\theta_{x}}$, and $\mathrm{\theta_{y}}$ axes were observed to be 11$\mathrm{nm}$, 644$\mathrm{nrad}$, and 650$\mathrm{nrad}$, respectively. On the other hand, with respect to the output displacement and rotations, there were within error of $\%0.045$ along the Z-axis, $\%0.088$ along the $\mathrm{\theta_{x}}$-axis and $\%0.148$ along the $\mathrm{\theta_{y}}$-axis, respectively.

\vspace{-0.5mm}
\subsection{Hysteresis Compensation}
Hysteresis is an undesired effect for precise manipulation using a piezo-actuated micromanipulator. Hence, a series of sinusoidal inputs were generated and applied to the developed micromanipulator to investigate the capability of the proposed control methodologies for hysteresis compensation. The results are presented in Figure \ref{FIG_hyst}. It can be observed that the hysteresis effect was compensated/eliminated utilizing the SMC-NDO control methodology.

Finally, the performances of micromanipulator and control methodologies regarding very complex motion tracking and hysteresis compensation are summarized in Table \ref{table_3}. According to this table, the maximum and minimum improvements of $\%$100 and $\%$67 were achieved, respectively, by implementing the SMC-NDO control methodology. The minimum bound of the improvement will increase if a smooth time-transitioned trajectory is considered to track, which is the case in many practical applications.

\vspace{-0.5mm}
\section{Conclusion}

A monolithic parallel $\mathrm{Z\theta_x\theta_y}$ micromanipulator with large range, high resolution, and high bandwidth frequency was presented. This positioning system utilized flexure-based components to produce desired motions in three out-of-plane directions. The linearized displacement relation between the actuation space and the output Cartesian space of the micromanipulator was established utilizing FEA. Hence, the micromanipulator's workspace was obtained implementing the inverse kinematics. Additionally, stress level in the flexure hinges was found to be below the yield point of the material (ABS) under the maximum input of the PEAs, thus verifying the repeatability and stability of the developed micromanipulator for the precise manipulation tasks. An SMC control methodology with nonlinear disturbance observer was introduced and utilized for the experimental evaluation of the micromanipulator. This control methodology was established to track very complex desired motion trajectories with high accuracy. It was also designed to accommodate system parametric uncertainties, cross-axis coupling, and nonlinearities including unknown disturbances and the hysteresis effect in the $\mathrm{Z\theta_x\theta_y}$ micromanipulator. The stability and robustness of the proposed closed-loop control technique (SMC-NDO) were proved, and the convergence of the position tracking errors to zero was guaranteed by the established system. Furthermore, an experimental facility was established to investigate the effectiveness of control methodologies and micromanipulator's high precision positioning capabilities. Based on the recorded experimental results, the positioning system showed very fine position and orientation resolutions of $\pm$4$\mathrm{nm}$, $\pm$250$\mathrm{nrad}$, and $\pm$230$\mathrm{nrad}$ throughout the micromanipulator's motion range of $\pm$238.5$\mathrm{\mu m}$ $\times$ $\pm$4830.5$\mathrm{\mu rad}$ $\times$ $\pm$5486.2$\mathrm{\mu rad}$. Finally, based on the root mean square error (RMSE) analysis of trajectory motion tracking and maximum hysteresis investigation, high resolution, high accuracy, hysteresis elimination, and consequently the effectiveness of the SMC-NDO control methodology were established.

\vspace{-2mm}

\section*{Acknowledgment}
This work was funded by the Australian Research Council (ARC) Discovery Project (DP) grant, and the Australian Research Council (ARC) Linkage Infrastructure, Equipment and Facilities (LIEF) grant.

\vspace{-2mm}

\section*{Conflict of interest}
The authors declare that they have no conflict of interest.

%\begin{spacing}{0.95}

%\section*{References}

% References

%\begin{thebibliography}{26} 

%\bibitem[22]{Ali5} M. Ghafarian, B. Shirinzadeh, A. Al-Jodah, T.K. Das, J. Pinskier, "Design and FEA based optimization of a monolithic z/tip/tilt mechanism", Manuscript submitted for publication.

\bibliographystyle{myIEEEtran}
\bibliography{library}

\vfill


%\vspace{-3mm}

%\begin{flushleft}
%\RaggedRight
%\justify
%\footnotesize{}	
%\end{flushleft}

%\end{IEEEbiography}

%\end{spacing}

%\end{thebibliography}

\end{document}